\documentclass{isprs} % isprs class modified 23-04-2019 (Dennis Wittich)
\usepackage{subfigure}
\usepackage{setspace}
\usepackage{geometry} % added 27-02-2014 Markus Englich
\usepackage{epstopdf}
\usepackage[labelsep=period]{caption}  % added 14-04-2016 Markus Englich - Recommendation by Sebastian Brocks
\usepackage[british]{babel} 
\usepackage[hang]{footmisc}
\usepackage{booktabs}
\usepackage{hyperref}
\usepackage{pifont}
\usepackage{multirow}
\usepackage{tabularx}
\usepackage{siunitx}
\usepackage{multirow}
\usepackage{glossaries}
\newacronym{LOD}{LoD}{Level of Detail}
\newacronym{MLS}{MLS}{Mobile Laser Scanning}
\newacronym{ALS}{ALS}{Airborne Laser Scanning}
\newacronym{GAN}{GAN}{Generative Adversarial Network}
\newacronym{FFC}{FFC}{Fast Fourier Convolution}
\newacronym{DM}{DM}{Diffusion Probabilistic Model}
\newacronym{SSIM}{SSIM}{Structural Similarity Index}
\newacronym{IoU}{IoU}{Intersection over Union}
\newacronym{LPIPS}{LPIPS}{Learned Perceptual Image Patch Similarity}
\newacronym{BIM}{BIM}{Building Information Modeling}
\newacronym{OGC}{OGC}{Open Geospatial Consortium}
\newacronym{GML}{GML}{Geography Markup Language}
\newacronym{XML}{XML}{Extensible Markup Language}
\newacronym{PDE}{PDE}{Partial Differential Equations}
\newacronym{SD}{SD}{Stable Diffusion}
\usepackage{todonotes}
\usepackage{pifont}

\usepackage{comment}

\begin{document}

\title{FacaDiffy: Inpainting unseen facade parts using diffusion models}
\date{}

% KAO: Remove extra spacing

% Anonymous submissions, authors' names should not be visible
% \author{
%  Orhan Altan\textsuperscript{1}, Ian Dowman\textsuperscript{2}, Florent Lafarge\textsuperscript{3}, Clément Mallet\textsuperscript{4}, Christian Heipke\textsuperscript{5} }
\author{Thomas Fröch\textsuperscript{$\ast$,1}, Olaf Wysocki\textsuperscript{2}, Yan Xia\thanks{Corresponding Author}\textsuperscript{~~,3,5} , Junyu Xie\textsuperscript{4}, Benedikt Schwab\textsuperscript{1}, Daniel Cremers\textsuperscript{3,5}, Thomas H. Kolbe\textsuperscript{1}}

% KAO: Remove extra newline

% Anonymous submissions, authors' affiliations should not be visible
\address{
	\textsuperscript{1}Chair of Geoinformatics, TUM School of Engineering and Design, Technical University of Munich (TUM),\\Munich, Germany - (thomas.froech, benedikt.schwab, thomas.kolbe)@tum.de\\ 
    \textsuperscript{2} Photogrammetry and Remote Sensing, TUM School of Engineering and Design, Technical University of Munich (TUM),\\Munich, Germany - olaf.wysocki@tum.de\\ 
	\textsuperscript{3}Computer Vision Group, TUM School of Computation, Information and Technology, Technical University of Munich (TUM),\\Munich, Germany - (yan.xia, cremers)@tum.de\\
    \textsuperscript{4}Visual Geometry Group, Department of Engineering Science, University of Oxford. Oxford, UK - jyx@robots.ox.ac.uk\\
    \textsuperscript{5}Munich Center for Machine Learning, Munich, Germany\\
	
}
%\address{**** (for review, affiliations must be rendered anonymous)}

% If the corresponding author is NOT the final author, always add a % space before the subsequent comma, i.e.
% first author name\textsuperscript{a,}\thanks{Corresponding author} , % second author name \textsuperscript{b}, etc.
% thanks to Niclas Borlin 05-05-2016
% information on the corresponding author should not be used any longer and has been commented out
% C. Heipke, Jan 03,2024

% the use of the information of commissions and working groups should not be used any longer and has been commented out
% C. Heipke, Sept. 20,2022
%\commission{XX, }{YY} %This field is optional. If filled, XX and YY should be replaced by adequate numbers. See https://www2.isprs.org/commissions/
%\workinggroup{XX/YY} %This field is optional.
%\icwg{}   %This field is optional.

% KAO: Use times symbol
\abstract{
High-detail semantic 3D building models are frequently utilized in robotics, geoinformatics, and computer vision. One key aspect of creating such models is employing 2D conflict maps that detect openings' locations in building facades. Yet, in reality, these maps are often incomplete due to obstacles encountered during laser scanning.
To address this challenge, we introduce FacaDiffy, a novel method for inpainting unseen facade parts by completing conflict maps with a personalized Stable Diffusion model. Specifically, we first propose a deterministic ray analysis approach to derive 2D conflict maps from existing 3D building models and corresponding laser scanning point clouds. Furthermore, we facilitate the inpainting of unseen facade objects into these 2D conflict maps by leveraging the potential of personalizing a Stable Diffusion model. To complement the scarcity of real-world training data, we also develop a scalable pipeline to produce synthetic conflict maps using random city model generators and annotated facade images. 
Extensive experiments demonstrate that FacaDiffy achieves state-of-the-art performance in conflict map completion compared to various inpainting baselines and increases the detection rate by $22\%$ when applying the completed conflict maps for high-definition 3D semantic building reconstruction.
The code is be publicly available in the corresponding GitHub repository: https://github.com/ThomasFroech/InpaintingofUnseenFacadeObjects
}


\maketitle

%\saythanks % added 28-02-2014 Markus Englich

\section{Introduction}\label{MANUSCRIPT}

Semantic 3D city models hold significant potential to address pressing global issues. 
Unlike mesh-based models, they are characterized by watertightness and object-oriented modeling, which has proven pivotal in various applications, such as estimating building solar potential and simulating wind flow~\cite{Biljecki2015}. 
Currently, they are ubiquitous, as, for example, approximately 140 million open access building models are available in the United States, Switzerland, and Poland while 55 million are available in Germany \cite{awesomeCityGMLPaper}.
% To represent, store, and exchange semantic 3D city models, the standard OGC CityGML defines a conceptual data model and exchange format \cite{kolbeOGCCityGeography2021}. 
While numerous cities and entire countries provide semantic 3D building models at \gls{LOD}2, characterized by complex roof shapes and planar facades, \gls{LOD}3 datasets with semantically detailed facades remain scarce.

However, such highly-detailed \gls{LOD}3 datasets are required for numerous application areas, such as assessing flood risk~\cite{amirebrahimi2016bim}, analyzing building potential for vertical farming~\cite{palliwal20213d}, and estimating energy demand ~\cite{nouvel2013citygml}. 

\begin{figure}[htpb]
      \centering
     \includegraphics[width=0.38\textwidth]{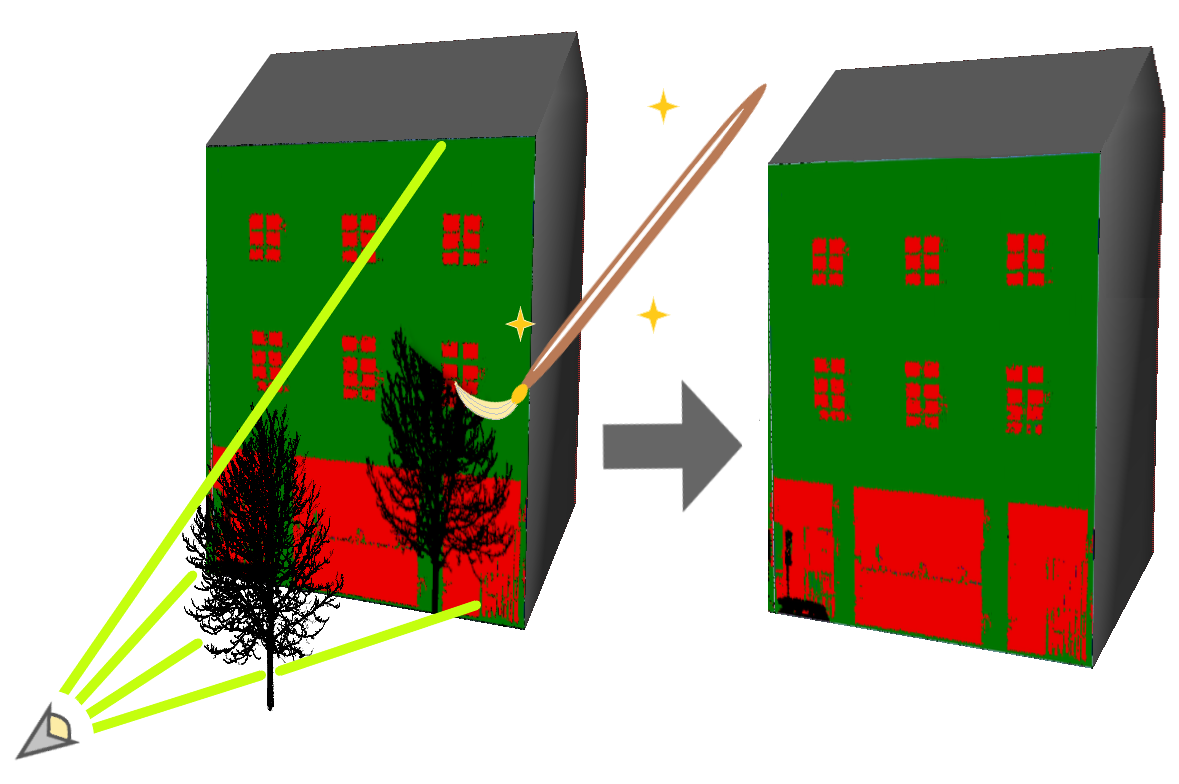}
     \caption{The conflict maps are prone to occlusions (black) owing to the reliance on the ray-to-model (yellow) analysis. We complement conflicts (red) and confirmations (green) by employing the proposed FacaDiffy.}
     % a deep-learning-based image inpainting strategy.}
     %\caption{Laser rays often encounter obstacles, such as vegetation, which obstructs their path and obscures building facades rendering the conflict maps incomplete.
     %We address this challenge by employing deep-learning-based image inpainting strategies.
     %This approach enables us to obtain completed 2D conflict maps.}
     \label{fig:funny_intro}
\end{figure}

Although a great deal of research has been devoted to the automatic \gls{LOD}3 reconstruction \cite{szeliski2010computer}, the current practice indicates that tedious, manual \gls{LOD}3 modeling prevails~\cite{manualLoD3seismic}. 
Yet, recent developments have shown that the so-called \textit{conflict maps} prove to be valuable for the automatic \gls{LOD}3 reconstruction~\cite{wysocki2023scan2lod3}. 
As we illustrate in \autoref{fig:funny_intro}, such conflict maps are generated based on the analysis of the 3D building model and sensor rays: The surface is deemed confirmed (green) when the ray point hits the surface, conflicted (red) when the ray traverses the surface, and unknown (black) when the surface is unmeasured.
These maps are considered a core intermediate geometric reconstruction cue, as they are frequently coupled with semantic-rich images or point clouds \cite{wysocki2023scan2lod3}; and the principle can also be directly used, for example, for building change detection \cite{tuttas2015validation}.
However, as we see in \autoref{fig:funny_intro}, the conflict maps are prone to occlusions (black), which render them incomplete and impact their at-scale applicability. 

Recent machine learning advancements have provided potent models for image inpainting, such as Stable Diffusion \cite{Rombach_2022_CVPR} or the LaMa GAN \cite{Suvorov2021}, presenting an opportunity to address the challenge of incomplete conflict maps.
We introduce the application of deep- learning methods to inpaint previously unseen facade objects into 2D conflict maps computed from existing \gls{LOD}2 models and corresponding point clouds. 
Completed conflict maps can, for example, be incorporated into existing pipelines for \gls{LOD}3 reconstruction and contribute to increasing their reconstruction accuracy or be utilized for change detection \cite{tuttas2015validation}.
We facilitate the deployment of a Stable Diffusion inpainting model \cite{Rombach_2022_CVPR} by personalizing with Dreambooth \cite{Ruiz2023}, utilizing synthetic conflict maps derived from randomly generated semantic city models and those obtained from annotated images of the CMP Facade Database \cite{Tylecek2013} as training data.

To summarize, our contributions are as follows:

\begin{itemize}
    \item A deterministic method to generate 2D conflict maps from existing \gls{LOD}2 building models and corresponding laser scanning point clouds.
    
    \item Conflict-oriented, personalized stable diffusion inpainting improving the conflict map completeness.
    
    \item An approach to generate synthetic conflict maps using \\ stochastic city model generators.
    
\end{itemize}
% KAO: Sloppy spacing ensures non-overfull lines. Can be removed if this is not an issue.
\sloppy

\section{Related Work}

\paragraph{Semantic 3D city models.}
In addition to offering geometric and visual insights into topographic features, semantic 3D city models provide comprehensive information about structures, taxonomies, and aggregations at the scale of cities, regions, and even complete countries. % \cite{Groeger2012}.
For the representation and management of city models, the standard CityGML is used internationally, which has been issued by the \gls{OGC} \cite{Kolbe2009,Groeger2012,kolbeOGCCityGeography2021}.
CityGML enables the modeling of urban objects
with their 3D geometry, appearance, topology, and semantics at four different \glspl{LOD}.
The data model of CityGML 3.0 is based on the ISO 191xx series of geographic information standards, and CityGML datasets can be encoded using the \gls{GML} \cite{Kutzner2020}.
%Its semantic model is based on the ISO 19100 standards family framework for modeling geographic features.
% CityGML, an \gls{OGC} standard since 2008 \cite{Groger2008,Kolbe2009}, facilitates the representation of buildings at various semantic \gls{LOD}.
%The implementation takes the form of an application schema for \gls{GML}, utilizing \gls{XML}.

\paragraph{Synthetic generation of semantic city models.}
With Random3Dcity, Biljecki \textit{et al.} \cite{BiljeckiRandom3Dcity} introduce a method for the procedural generation of randomized semantic city models at various \gls{LOD} levels.
Their approach utilizes a set of pre-defined architectural modeling rules guiding its stochastic nature.
Such rules govern aspects like the permissible positioning of facade elements such as doors or windows.

\paragraph{Reconstruction of semantic 3D building models.}
The considerable potential for applying detailed \gls{LOD}3 models across diverse domains, coupled with their scarcity, has motivated a significant number of studies to explore the reconstruction of such models.
Investigations involve leveraging various data sources, such as optical images, oblique \gls{ALS} point clouds, and \gls{MLS} measurements, as well as employing diverse approaches, including formal grammar approaches and Bayesian networks \cite{ripperda_rekonstruktion_2010,helmutMayerLoD3,wysocki2023scan2lod3}.
Within their pipeline for reconstructing underpasses in semantic \gls{LOD}2 city models from co-registered \gls{MLS} point clouds, Wysocki \textit{et al.} introduce the concept of 2D conflict maps. 
Their probabilistic approach relies on an occupancy grid implemented as an octree structure, with voxel sizes reflecting the combined uncertainty of the \gls{MLS} measurements and the semantic city model.
%\olafworries{
This concept has been further developed by \cite{Wysocki_Conflict_Map,wysocki2023scan2lod3,hoegner2022automatic}, thereby substantiating its effectiveness.
%}
%\olafworries{
The pivotal advantage of the previously mentioned methods over mesh-based approaches is the usage of 3D semantic building models as priors, which has proven to maintain the model watertightness as well as higher 3D reconstruction accuracy, reaching up to around $50\%$ when compared to the mesh-based Poisson reconstruction~\cite{wysocki2023scan2lod3}. 
Nevertheless, despite generally yielding commendable results, the challenge of incompleteness remains unsolved.
\paragraph{Deep-learning-based image inpainting.}
% Besides traditional image inpainting methods that often rely on solving \gls{PDE}s \cite{Telea2004,Bertalmio2001}, recent advancements in the dynamic research domain of image inpainting provide an opportunity to address the challenge of incomplete 2D conflict maps. 
Besides traditional image inpainting methods, which typically rely on solving \gls{PDE}s \cite{Telea2004,Bertalmio2001}, and yield unsatisfactory results when applied to facade images \cite{fritzsche2022inpainting}. 
Recent advancements in deep-learning-based image inpainting suggest potential methodologies for addressing the challenge of incomplete 2D conflict maps. 
Large mask inpainting (LaMa) \cite{Suvorov2021}, configured as a Generative Adversarial Network (GAN), employs Fast Fourier Convolution operators \cite{Chi2020FFC} to overcome the limitation of restricted receptive fields, thereby enabling expansive coverage across the entire image.

% Large mask inpainting (LaMa) \cite{Suvorov2021} is configured as a Generative Adversarial Network (GAN) following a structure similar to a feed-forward ResNet inpainting network.
% To overcome the constraint of a restricted receptive field, LaMa employs Fast Fourier Convolution operators \cite{Chi2020FFC}, enabling an expansive coverage across the entire image size.

% \gls{DM}s such as \gls{SD} \cite{Rombach_2022_CVPR} represent powerful methods for various applications including image inpainting.
% %By leveraging a mapping into a latent feature space, Rombach \textit{et al.} significantly reduce the required computational resources for the deployment of \gls{DM}s.
% \gls{SD} demonstrates remarkable flexibility by utilizing conditioning mechanisms and cross-attention layers in the architecture to integrate additional information.
% Incorporating five specialized input channels combined with a dedicated training procedure enables its application for image inpainting tasks.
% %Notably, unlike many inpainting strategies, 
% The \gls{SD} model requires the provision of a text prompt for image inpainting.

\gls{DM}s have emerged as powerful tools for various generative applications in the last years. 
Notably, \gls{SD} \cite{Rombach_2022_CVPR} demonstrates remarkable flexibility with its capability to handle open-ended text conditioning. When supplemented with additional image and mask inputs, the \gls{SD} framework can be further extended to solve image inpainting tasks, guided by relevant text prompts. 
To our knowledge, no method deploying diffusion models for inpainting facade conflict maps has been published yet.

To tailor pre-trained \gls{SD} models for domain-specific applications, a variety of personalization ({\em i.e.,} customization) techniques \cite{gal2022textual,kumari2022customdiffusion,Wei_2023_ICCV} are developed. These methods generally utilize small-scale datasets to fine-tune the pre-trained \gls{SD} pipeline for specialized domains. In particular, Dreambooth \cite{Ruiz2023} stands out as a robust personalization approach, incorporating a prior-preservation loss and LoRA~\cite{hu2022lora}-based fine-tuning. 
%This method is further proven to be effective for the inpainting variant of \gls{SD} \cite{von-platen-etal-2022-diffusers}. 
% Dreambooth \cite{Ruiz2023} is a method designed for the personalization of large text-to-image diffusion models.
% It facilitates subject-driven fine-tuning, enabling the network to generate a range of new examples for a learned instance of an object type.
% A unique identifier, linked to the specific instance, is employed in the text-prompt to reference the learned instance during inference.
% %A significant advantage of Dreambooth is the small number of training samples.
% Besides text-to-image applications, Dreambooth can also be applied to the image inpainting variant of \gls{SD} \cite{von-platen-etal-2022-diffusers}.

\section{Method}
\begin{figure*}[htpb]
     \centering
     \includegraphics[width=0.9\textwidth]{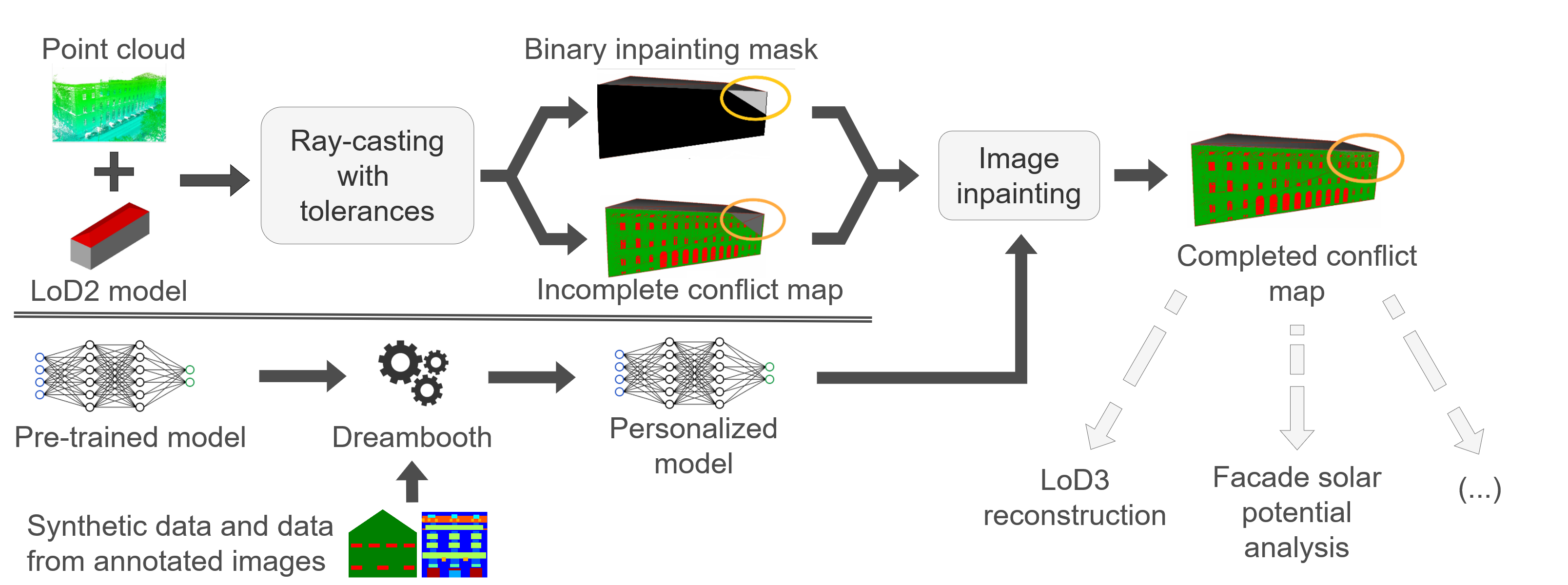}
     \caption{Schematic overview of FacaDiffy. By combining an existing LoD2 building model and corresponding laser scanning point clouds, we formulate a deterministic method based on ray-casting analysis to obtain incomplete conflict maps and corresponding binary inpainting masks (top branch). We generate synthetic conflict maps to personalize the Stable Diffusion model (bottom branch), which is employed for inpainting given the partial evidence of deterministic conflict maps. These can be utilized for various downstream applications such as accurate LoD3 reconstructions, facade solar potential analysis, etc}
     \label{fig:overall-concept}
 \end{figure*}

As illustrated in \autoref{fig:overall-concept}, given the \gls{LOD}2 model and point cloud data, FacaDiffy aims at completing facade conflict maps by leveraging deep-learning-based image inpainting techniques. 
Specifically, Sect.~\ref{subsec:conflict-map} introduces a ray-casting approach to derive 2D real-world conflict maps and corresponding binary masks indicating occluded regions. These estimated occluded areas are then adopted as inpainting masks.
% Sect.~\ref{subsec:synthetic-conflict-map} elaborates on the method for ray-analysis-free synthetic conflict map generation
To complement insufficient real-world conflict map data for training, Sect.~\ref{subsec:synthetic-conflict-map} details a scalable pipeline that generates synthetic conflict maps as additional training data for the inpainting model. 
Sect.~\ref{subsec:inpainting} elaborates on a personalized \gls{SD}-based inpainting method designed to recover complete conflict maps.

\subsection{Conflict Map Computation}
% \subsection{Real-world Conflict Map Prediction}
\label{subsec:conflict-map}

As Figure \ref{fig:overall-concept} illustrates, we obtain incomplete conflict maps and corresponding binary masks, highlighting missing ({\em i.e.,} occluded) areas through a deterministic ray-casting approach with tolerances that combine semantic \gls{LOD}2 building models and corresponding laser scanning point cloud data.

We first define the unit rays, denoted as $\textbf{r}_\textbf{p}$, as originating from a viewpoint $\textbf{v}$ and oriented towards a corresponding point $\mathbf{p}$.

\vspace{-0.3cm}

\begin{equation}\label{eq:RayDefinition}
    \mathbf{r}_{\mathbf{p}} = \mathbf{v}+\frac{\mathbf{p}-\mathbf{v}}{|\mathbf{p}-\mathbf{v}|}
\end{equation}

\begin{figure}[htpb]
    \centering
     \includegraphics[width=0.45\textwidth]{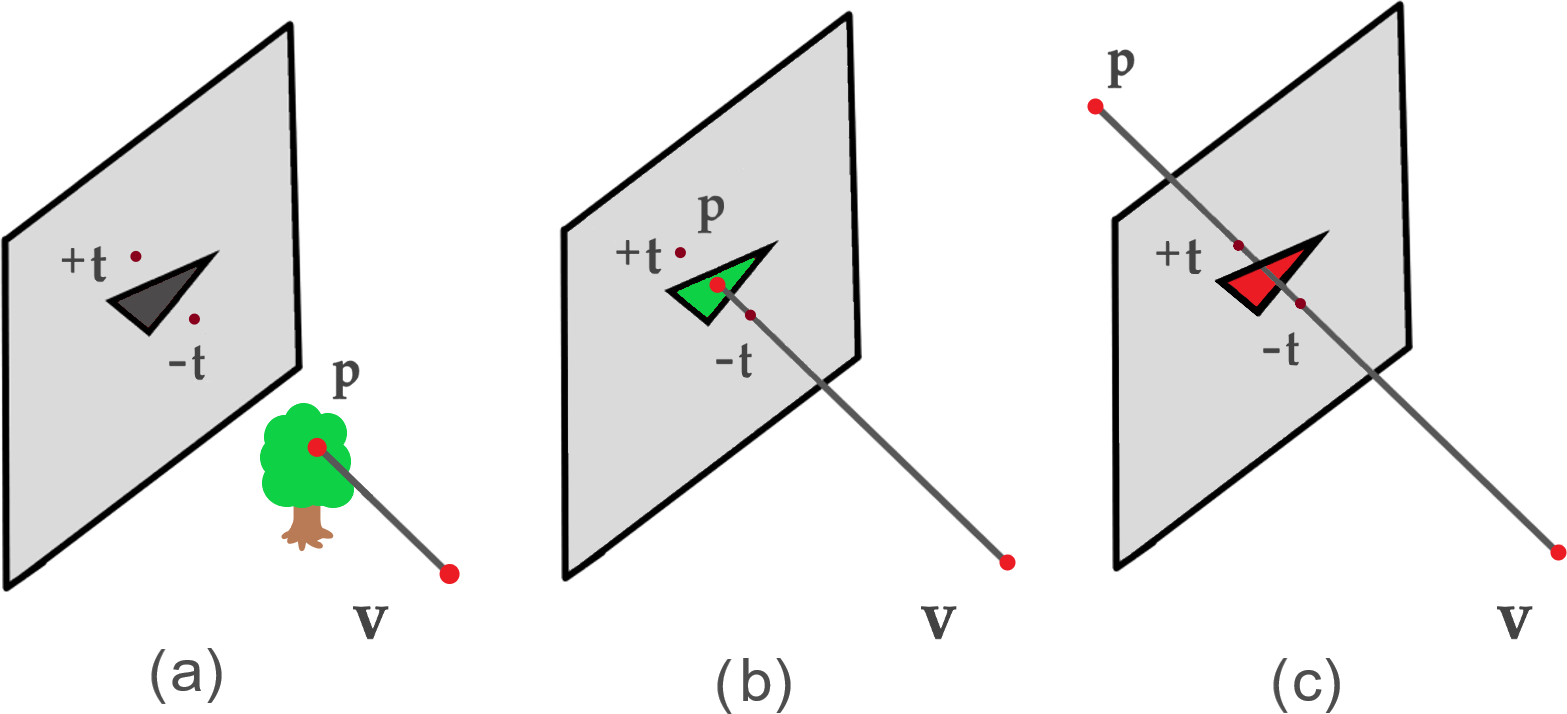}
     \caption{Schematic overview of the conflict determination in the ray casting approach with the viewpoint $\mathbf{v}$, the point $\mathbf{p}$, and the tolerance $\pm t$. Three distinct scenarios are illustrated: (a) unknown; (b) confirming; (c) conflicting.}
     \label{fig:ConflictDetermination}
\end{figure}

Then, we identify conflicts between a \gls{LOD}2 model and a corresponding point cloud by evaluating the distance to the intersection of the \gls{LOD}2 surface and the rays. 
Three mutually exclusive cases have to be distinguished: 
\begin{itemize}
\item Unknown: As illustrated in Figure \ref{fig:ConflictDetermination} (a), occurrences of occlusions caused by objects, such as vegetation, which are unrelated to a facade, are identifiable by an intersection distance shorter than the anticipated value.
This information is utilized to derive the binary masks indicating the {occluded areas} in the conflict maps.
\item Confirming: The majority of instances corresponds to Figure \ref{fig:ConflictDetermination} (b), where the intersection distance remains within an expected tolerance range $[-t, t]$, leading to {incomplete conflict map} predictions.
% , considering a tolerance denoted as $\pm \text{t}$. 
\item Conflicting: As depicted in Figure \ref{fig:ConflictDetermination} (c), the assessed intersection distance surpasses the expected value for openings in the facade, such as windows, due to the voyeur effect \cite{Tuttas2013}, or underpasses and arches that are not considered in \gls{LOD}2 models \cite{Kutzner2020}.
\end{itemize}

% Three cases have to be distinguished.
% As illustrated in Figure \ref{fig:ConflictDetermination} (a), occurrences of occlusions caused by objects, such as vegetation, which are unrelated to a facade, are identifiable by an intersection distance shorter than the anticipated value.
% This information is utilized to derive the binary masks indicating the \textbf{occluded areas} in the conflict maps.
% The majority of instances corresponds to Figure \ref{fig:ConflictDetermination} (b), where the intersection distance remains within an expected tolerance range $[-t, t]$, leading to \textbf{incomplete conflict map} predictions.
% % , considering a tolerance denoted as $\pm \text{t}$. 
% In the third case, depicted in Figure \ref{fig:ConflictDetermination} (c), the assessed intersection distance surpasses the expected value for openings in the facade, such as windows, due to the voyeur effect \cite{Tuttas2013}.

%In the third case, illustrated in Figure \ref{fig:ConflictDetermination} (c), owing to the voyeur effect \cite{Tuttas2013}, the assessed intersection distance exceeds the expected value for openings in the facade, such as windows.

%\olafworries{please name the three states here explicitly and explain - use your Fig3 as the reference}

To achieve a high geometric resolution, we employ a multi-iteration midpoint triangle subdivision approach \cite{chen2014general} on the triangles comprising the wall surface.
As the number of triangle subdivision iterations, denoted as $n_\text{div}$, increases, striking a balance between geometric resolution and computational efficiency becomes essential.

\subsection{Synthetic Conflict Map Generation}
% \subsection{Synthetic Conflict Map Generation}
\label{subsec:synthetic-conflict-map}
%\olafworries{We also introduce model-based deterministic conflict map generation.}
We aim to utilize realistic synthetic conflict maps to complement insufficient real-world training data.
These maps are derived from randomly generated semantic building models and classified facade image benchmarks, by leveraging their structured knowledge along with insights from previous works.
%We derive synthetic conflict maps from randomly generated semantic building models and classified facade image benchmarks by leveraging their provided structured knowledge and previous works' findings.
%\olafworries{We obtain the models .... and images ... }
%that we obtain by employing the Random3Dcity method \cite{BiljeckiRandom3Dcity} to create randomly generated semantic building models.

The extensive semantic information contained in the randomly generated CityGML \gls{LOD}3 models makes it possible to classify structures as conflicting (red) or confirming (green), according to prior knowledge about the behavior of certain types of building parts.
Windows or doors are considered to be conflicting due to the voyeur effect, while underpasses and extruded facade objects such as balconies or decorative molding deviate from the wall surface geometrically  \cite{Tuttas2013,tuttas2015validation,hoegner2022automatic,Wysocki_Conflict_Map}.
The same objects, identifiable by their annotation, are considered conflicting in the annotated facade images.

We subsequently project the facades to 2D and plot the projected triangles they comprise in the corresponding color to obtain synthetic 2D conflict maps.
%By subsequent facade-wise projecting and plotting of the triangles the randomly generated facades comprise, we obtain synthetic 2D conflict maps.
We apply these for personalizing with Dreambooth and to evaluate the inpainting performance in scenarios that closely match real-world applications.
%\olafworries{can you describe here your approach to the synthetic maps? You reformulate the conflict map problem here, so it might be quite interesting and is the core part of the method (it won't work without it)}
%\olafworries{it reformulates 'cause we don't use any rays there anymore. Meaning - the conflict can be more than ray-to-model analysis, actually}
\subsection{Deep-Learning-Based Inpainting}
\label{subsec:inpainting}
Given the incomplete conflict maps, we leverage the \gls{SD}-inpainting method~\cite{Rombach_2022_CVPR} to recover the missing areas. Specifically, we treat the estimated occluded regions as inpainting masks ({\em i.e.,} areas where the inpainting is conducted). 
% Additionally, as a pre-processing step, we convert the color-coded conflict maps into binary formats. This prevents potential color-associated biases during the inpainting process. 
We convert color-coded conflict maps into binary images, indicating the presence or absence of conflicts, before initiating the inpainting process to avoid undesirable structures related to color properties during inpainting.
However, directly applying the pre-trained \gls{SD}-inpainting may result in undesired artifacts, largely due to the domain discrepancy between the binary conflict maps and the real-world image priors embedded in the pre-trained \gls{SD} model. To mitigate such disparity, we adopt a personalization approach and carefully design the input text prompts.
% Once incomplete conflict maps and corresponding occluded regions are obtained, we treat the latter as the masked area and recover the correct conflict maps through image inpainting.
% Specifically, we apply the \gls{SD}-inpainting model as the major inpainting method.
% However, directly applying the \gls{SD}-inpainting model could lead to unexpected artifacts in the inpainted results, owing to the domain gaps between the conflict maps ({\em i.e.,} binary images) and the real-world images on which the \gls{SD}-inpainting model was pre-trained. 

% This conversion is then reversed once the inpainting is completed.
% \paragraph{Personalizing with Dreambooth}

\paragraph{Personalization of the inpainting model.} 
To adapt the pre-trained \gls{SD}-inpainting model for conflict map inpainting, we utilize the Dreambooth~\cite{Ruiz2023} technique.
% During the personalization fine-tuning, we consider both synthetically generated conflict maps and those derived from annotated images as training images.
During the personalization fine-tuning, we adopt the synthetically generated conflict maps (detailed in Sect. \ref{subsec:synthetic-conflict-map}) as training images, and apply random inpainting masks generated by the method proposed in \cite{von-platen-etal-2022-diffusers}.
% The inpainting masks are randomly generated by the method proposed in \cite{von-platen-etal-2022-diffusers}.
% We utilize the Dreambooth implementation for the \gls{SD}-inpainting model that is available in the Diffusers GitHub repository.
% This personalization process for the \gls{SD} inpainting models relies on randomly generated masks \cite{von-platen-etal-2022-diffusers}.
% Since Dreambooth has proven to be sensitive towards the setting of hyperparameters \cite{Patil2022}, we ensure the comparability of our experiments by consistently deploying the same set of hyperparameter settings thoroughly.

% \begin{itemize}
% \item Synthetic conflict maps  
% \item Conflict maps derived from annotated facade images
% \end{itemize}

\paragraph{The choice of text prompts.}
% \junyucomment{}
% As a diffusion-based text-to-image model, Stable Diffusion \cite{Rombach_2022_CVPR} requires users to provide a text prompt for inference.
During the inpainting process, the choice of text prompts exhibits a great influence on resultant quality.
% The text-prompt exhibits great influence on the content that is inpainted into the missing areas.
To heuristically identify a suitable text-prompt that is consistently applied for conflict map inpainting, we investigate a variety of text-prompts, involving high-level ({\em e.g., ``Window''}) and low-level ({\em e.g., ``Rectangle''} ) descriptions and assessing their corresponding effects on the inpainting outcomes. 

% By investigating different text-prompts that focus on either high-level low low-level descriptions of the desired content to be inpainted, we identify a suitable text-prompt that we consistently apply.
%We aim to substitute the manual text prompt by automatically combining a set of pre-defined strings according to high-level properties of the images.
%We analyze the histogram, axis symmetry, and the fragmentation of the images to combine the pre-defined strings accordingly.
%This approach allows for some flexibility while maintaining a strong level of control over the resulting text-prompts.
\section{Experiments}
We provide a detailed discussion with additional examples, implementation details, and systematic tests in the supplementary material in the corresponding GitHub Repository\footnote{https://github.com/ThomasFroech/InpaintingofUnseenFacadeObjects}. 
\subsection{Datasets}
\paragraph{Real-world data for conflict maps.}
As sources for computing real-world conflict maps, we leveraged two primary datasets: (i) a proprietary MLS point cloud \cite{Wysocki2023MLS2LOD3}, acquired by the company 3D Mapping Solutions \cite{Haigermoser2015,mofa} with its geo-referencing supported by the German SAPOS RTK system; (ii) the official LoD2 building models supplied by the Bavarian State Office for Digitizing, Broadband and Survey\cite{BayerischeVermessungsverwaltung2023}.
%The German SAPOS RTK system was utilized in conjunction with a proprietary mobile mapping platform to realize the georeferencing of the point cloud.
The datasets we employed encompass sections of the TUM city campus and of the Pfisterstraße in Munich, Germany.

\paragraph{Annotated facade images.}
To derive conflict maps from annotated images, we utilized the CMP database of annotated images provided by the Center for Machine Perception in Prague \cite{Tylecek2013}.
We considered windows, doors, cornices, sills, balconies, blinds, decorations, molding, pillars, and background as conflicting.

\paragraph{Ground-truth information from LoD3 models.} We obtained ground-truth conflict maps from existing LoD3 building models maintained in the Tum2Twin GitHub repository \cite{TumToTwin}
\footnote{https://tum2t.win/}.
Note that, the limited availability of LoD3 building models with corresponding point clouds only supports a small-scale evaluation.

\paragraph{Inpainting masks.}
We consider two types of inpainting masks: the vegetation ({\em i.e.,} tree-shaped) mask and randomly generated masks. 
To produce realistic, tree-shaped binary masks, we utilized a point cloud stemming the TreeML-Data collection \cite{Yazdi2024}, projecting it to a 2D raster orthogonally. 
To produce medium-sized, randomly generated masks, we applied the method introduced by \cite{Suvorov2021}.

\subsection{Evaluation Metrics}
We assessed the similarity between the inpainting results and ground-truth information from two major perspectives: (i) The structural and shape similarity is measured by the Structural Similarity Index (SSIM) (sliding window size of $71$ pixels) \cite{Wang2004} and the Intersection over Union (IoU) \cite{Rezatofighi2019};
(ii) Given the limitations of IoU in accurately assessing semantic similarity (\autoref{tab:Results1}, \autoref{tab:ResultsAblation}), we also employ the Learned Perceptual Image Patch Similarity (LPIPS) \cite{Zhang2018} for perceptual assessment.

% \subsection{Experimental Results}
\subsection{Implementation Parameters}
\begin{figure*}[htpb]
     \centering
     \includegraphics[width=0.9\textwidth]{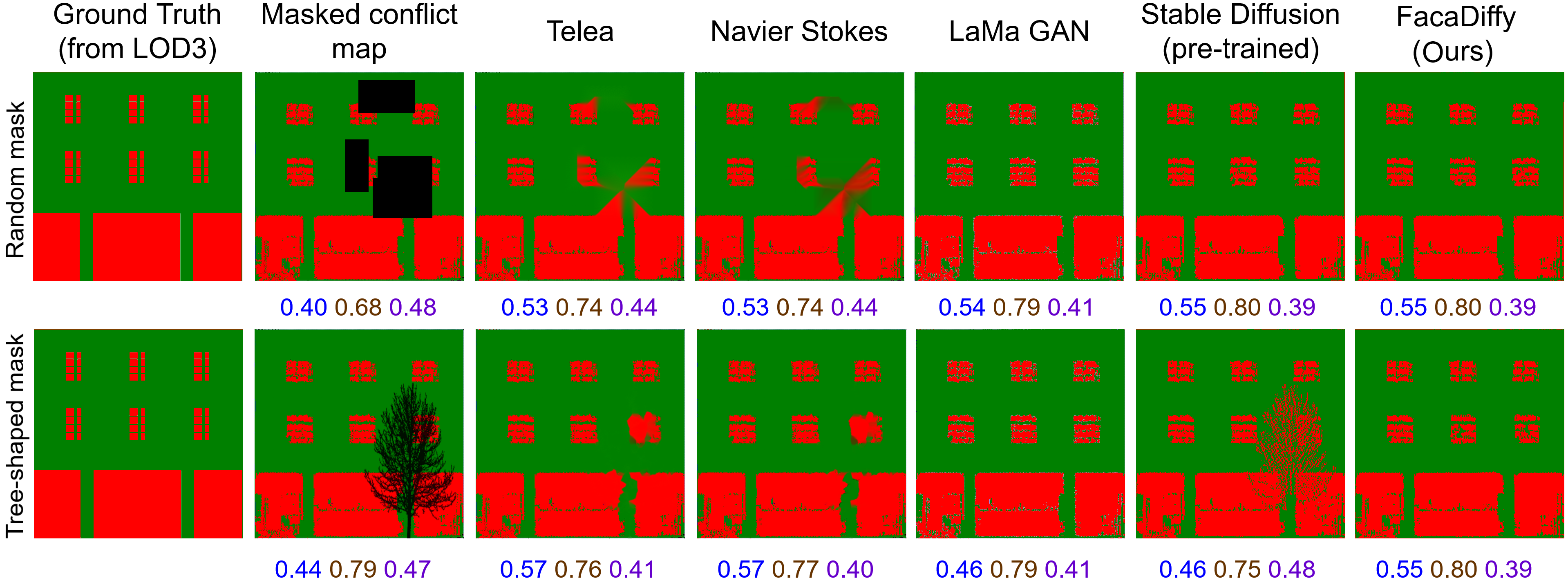}
     \caption{Exemplary inpainting results on a real conflict map. The similarity between ground-truth (LoD3) and inpainted results is measured in terms of SSIM (blue), IoU (brown), and LPIPS (purple).
     The conflict maps are color-coded with the conflicting (red), confirming (green), and unknown/masked (black) areas.
     % The conflict maps were converted to black and white for the inpainting and were manually color-coded afterward.
     % Red: Conflicting, Green: Confirming, Black: Masked area.
     }
     \label{fig:Exemplary tree results}
     %\vspace{-0.5cm}
 \end{figure*}
\paragraph{Personalization with DreamBooth.}
% We utilize the DreamBooth implementation for the SD-inpainting model that is available in the Diffusers GitHub repository.
%\noindent\textbf{Conflict map computation.}
In our implementation of the deterministic ray analysis approach to generate conflict maps we set $t=0.7\si{\metre}$ and $n_\text{div}=8$.
We personalized the SD-inpainting model (sd-v1.5-inpaint) with DreamBooth leveraging $192$ synthetically generated conflict maps that are derived from randomly generated semantic building models obtained from the Random3Dcity application \cite{BiljeckiRandom3Dcity}. 
Following the implementation of \cite{inpaitning_fine_tuning2024}, inpainting masks were randomly generated following \cite{von-platen-etal-2022-diffusers}. 
We specified the text prompt as \textit{``Black background with 
white patches that are consistent and symmetric to the rest of the image''}.
In choosing hyperparameters, we followed the settings specified in \cite{Patil2022} and consistently deployed them throughout all experiments. 

\paragraph{LaMa GAN.}
As an alternative deep-learning-based inpainting strategy, we trained the LaMa GAN \cite{Suvorov2021} for $25$ epochs.
%Utilizing our approach discussed in section \ref{subsec:synthetic-conflict-map}, 
We leveraged the CMP-base dataset \cite{Tylecek2013} and the Random3Dcity application \cite{BiljeckiRandom3Dcity} to obtain a training dataset consisting of approx. $20.000$ conflict maps.

% \smallskip
\paragraph{Traditional inpainting strategies.}
For comparison with traditional inpainting strategies based on solving Partial Differential Equations (PDE)s, we applied the inpainting method by Telea \cite{Telea2004} and Navier-Stokes-based inpainting \cite{Bertalmio2001}.
%to conflict maps derived from the CMP-extended database \cite{Tylecek2013}.
We utilized the implementation available in \cite{opencv_library}.
Evaluation results of both traditional methods are summarized in \autoref{tab:Results1}.

\subsection{Comparisons with State-of-the-art}
We evaluated our inpainting model performance on two major datasets with (i) $228$ conflict maps derived from the annotated facade images in the CMP-extended database \cite{Tylecek2013}
and (ii) a small-scale dataset with real conflict maps concerning ground-truth data from corresponding \gls{LOD}3 building models, as a demonstration of our real-world applicability.
We treated the unmasked conflict maps as the ground truth and measured their similarity with the completed conflict maps as an indication of the inpainting quality.

\begin{table*}[t]
    \centering
    %\small
    \begin{tabular}{lrrrrrr}
       \toprule
        \multirow{2}{*}{Methods} & \multicolumn{3}{c}{Randomly generated masks}  & \multicolumn{3}{c}{Tree-shaped masks} \\
        \cmidrule(r){2-4}
        \cmidrule(r){5-7}
         & $\text{SSIM} \uparrow$ & $\text{IoU} \uparrow$ & $\text{LPIPS} \downarrow$ & $\text{SSIM} \uparrow$ & $\text{IoU} \uparrow$ & $\text{LPIPS} \downarrow$ \\
         \midrule
	      Masked conflict map & $0.72$ & $0.85$ & $0.24$ & $0.83$  & $0.92$ & $0.21$\\
        \midrule
        Telea \cite{Telea2004} & $0.82$ & $0.89$ & $0.18$ & $0.94$ & $0.96$ & $0.10$\\
        Navier-Stokes \cite{Bertalmio2001} & $0.85$ & $0.92$ & $0.14$ & $0.94$ & $0.97$  & $0.09$\\
        LaMa GAN \cite{Suvorov2021} & $0.85$ & $0.92$  & $0.14$ & $0.95$  & $0.94$ & $0.06$ \\
        SD-inpainting (pre-trained) \cite{Rombach_2022_CVPR} & $0.89$ & $0.72$ & $0.09$ & $0.85$ & $0.70$  & $0.21$\\
        FacaDiffy (Ours) & $0.91$ & $0.72$ & $0.08$ & $0.90$ & $0.66$  & $0.11$\\
        \bottomrule
    \end{tabular}
    \caption{Quantitative comparison with baseline methods. The evaluation is conducted on the CMP-extended image database (comprising 228 annotated images).
    The similarity is measured between the inpainting results and unmasked ground-truth conflict maps. The IoU metric yields counterintuitive results when applied to traditional inpainting strategies, despite evident semantic inconsistencies, as exemplified in \autoref{fig:Exemplary tree results}}
    
    \label{tab:Results1}
\end{table*}

\begin{table*}[htpb]
   \centering
    %\small
    \vspace{0.2cm}
    \begin{tabular}{cccrrrrrr}
       \toprule
         \multirow[2]{2}{*}{Exp.} & \multirow{2}{*}{\shortstack{No. of \\ conflict maps}} & \multirow{2}{*}{\shortstack{Type of \\ conflict maps}} & \multicolumn{3}{c}{Randomly generated masks}  & \multicolumn{3}{c}{Tree-shaped masks} \\
        \cmidrule(r){4-6}
        \cmidrule(r){7-9}
          & & &  $\text{SSIM} \uparrow$ & $\text{IoU} \uparrow$ & $\text{LPIPS} \downarrow$ & $\text{SSIM} \uparrow$ & $\text{IoU} \uparrow$ & $\text{LPIPS} \downarrow$ \\
        % \midrule
	% Masked conflict map & $0.72$ & $0.85$ & $0.24$ & $0.83$  & $0.92$ & $0.21$\\
        \midrule
        % A (pre-trained SD-inpainting) & - & - & $0.89$ & $0.72$ & $0.09$ & $0.85$ & $0.70$  & $0.21$\\
        A & 5 & Syn. & $0.90$ & $0.73$ & $0.08$ & $0.86$  & $0.66$  & $0.18$\\
        B & 228 & Real & $0.91$ & $0.70$ & $0.08$ & $0.84$ & $0.65$  & $0.22$ \\
        \midrule
        C (default) & 192 & Syn. & $0.91$ & $0.72$ & $0.08$ & $0.90$ & $0.66$  & $0.11$\\
		\bottomrule
    \end{tabular}
    \caption{Quantitative comparison with baseline methods. The evaluation is conducted on the CMP-extended image database (comprising 228 annotated images).
    The similarity is measured between the inpainting results and unmasked ground-truth conflict maps. The IoU metric yields counterintuitive results when applied to traditional inpainting strategies, despite evident semantic inconsistencies, as exemplified in \autoref{fig:Exemplary tree results}}
    \label{tab:ResultsAblation}
\end{table*}

% \subsection{Comparison with Other Baselines}
\paragraph{Comparison with LaMa GAN.}
Results shown in \autoref{tab:Results1} and \autoref{fig:Exemplary tree results} suggest that LaMa GAN excels at recovering fine structures such as tree-shaped masks, while our method also yields competitive results. 
In terms of randomly generated masks, FacaDiffy exhibits superior performance compared to LaMa GAN. These findings suggest the better applicability of FacaDiffy considering the scenario with large occlusions.

\begin{figure}[htpb]
    \centering
    %\vspace{-1.5cm} 
    \includegraphics[width=0.38\textwidth]{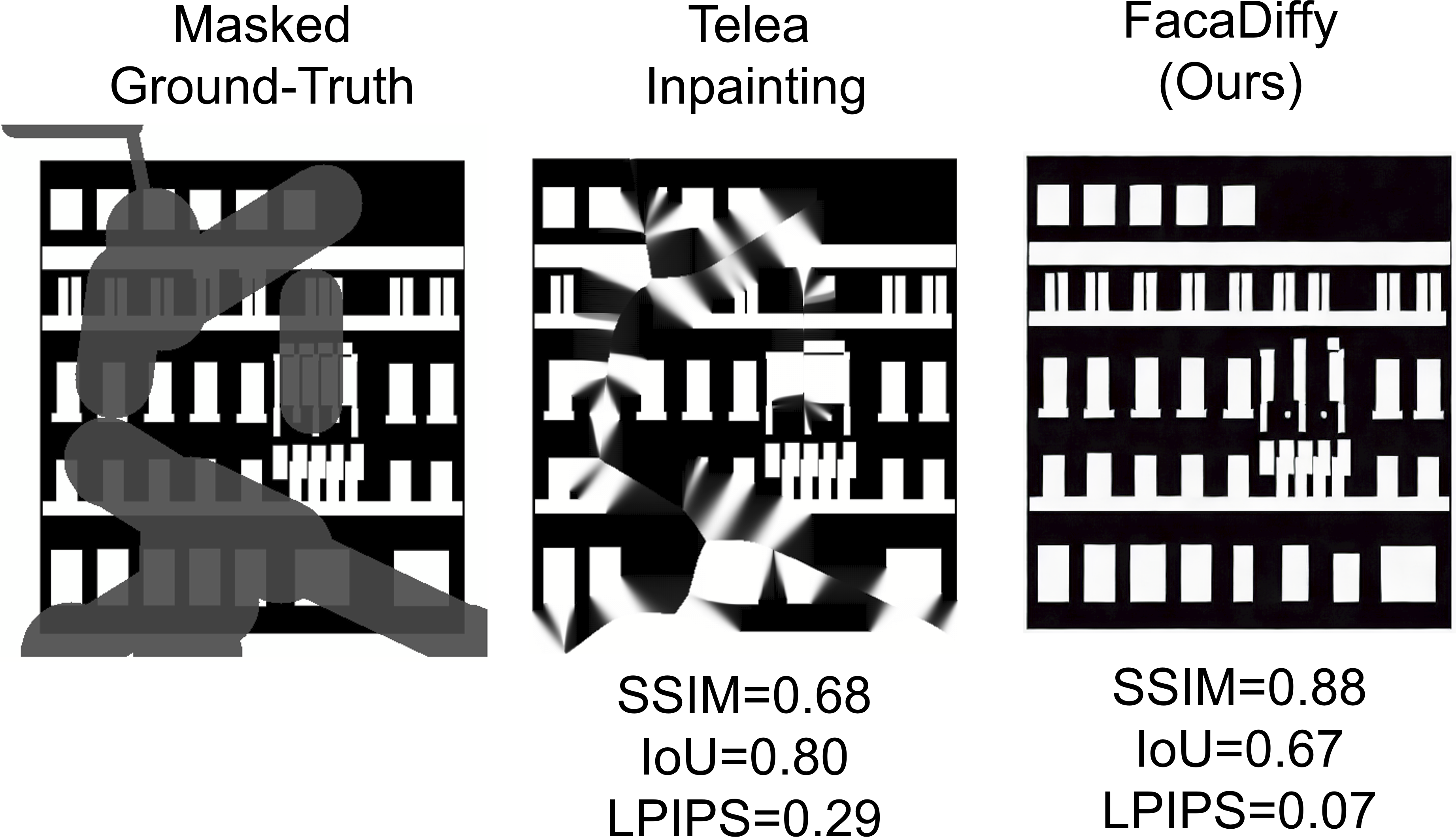}
    \caption{Counterintuitive IoU evaluation result for a randomly masked conflict map derived from the CMP-database of annotated images. While our method performs better qualitatively, the IoU yields counterintuitive results. }
    \label{fig:counterintuitive_Evaluation}
    %\vspace{-0.5cm} 
\end{figure}

\paragraph{Comparison with traditional inpainting strategies.} Concerning randomly generated masks, the diminished performance of the traditional methods compared to FacaDiffy becomes evident in the LPIPS and SSIM measurements summarized in  \autoref{tab:Results1}. 
According to \autoref{tab:Results1}, the traditional approaches perform better in the case of the tree-shaped mask. 
However, \autoref{fig:Exemplary tree results} illustrates notable semantic inconsistencies in the inpainting outcomes achieved through these methods, while FacaDiffy demonstrates qualitatively superior semantic consistency.

\paragraph{Counterintuitive IoU evaluation results.}
When deploying the baseline methods, greater overlap due to larger continuous areas being inpainted into the images may positively impact the IoU, even though the underlying true semantic similarity would not be positively affected. 
Such counterintuitive results, evident in \autoref{tab:Results1} and \autoref{tab:ResultsAblation}, are exemplarily illustrated in \autoref{fig:counterintuitive_Evaluation}.
The IoU contradicts the qualitative similarity assessment of the inpainting results to the ground truth, the SSIM, and the LPIPS measurements. 
While these metrics indicate greater similarity, the IoU suggests the opposite.
In contrast to evaluating the IoU instance-wise with respect to the individual facade features, we evaluate the entire conflict maps, which might also affect the score. 
Additionally, as \cite{Zhang2018} mention, no universal metric for quantifying image completion is available, which motivates us to consider multiple metrics for an unbiased comparison.

\subsection{Ablation Studies}
%Additionally to the personalization with $192$ synthetic conflict maps, we conducted personalizations involving $5$ synthetically generated conflict maps, and $228$ conflict maps derived from annotated facade images.
Additionally, we conducted personalizations involving $5$ synthetically generated conflict maps, and $228$ conflict maps derived from annotated facade images.
We ensured comparability by applying the same text prompt consistently.
An improvement of $-0.07$ in the LPIPS measurements concerning tree-shaped masks in \autoref{tab:ResultsAblation} demonstrates that increasing the number of synthetic training samples from $5$ (Exp.\,A) to $192$ (Exp.\,C) generally leads to an enhanced inpainting performance. The personalized model, fine-tuned on synthetic data (Exp.,C), competes well with its counterpart trained on real-world conflict maps (Exp.,B), validating our synthetic map generation pipeline (\autoref{tab:ResultsAblation}). %This is particularly evident in the tree-shaped mask scenario, where we observe an improvement of $-0.07$ in the LPIPS measurements.
% As demonstrated in Table \ref{tab:ResultsAblation}, the personalized model, when fine-tuned on synthetically generated data (Exp.\,C), achieves competitive performance compared to its counterpart fine-tuned on real-world conflict maps derived from annotation facade images (Exp.\,B). This outcome further validates the efficacy and reliability of our synthetic conflict map generation pipeline.

% Besides randomly generated inpainting masks, we also consider the tree-shaped mask during evaluation.
%
%When FacaDiffy is adopted for subsequent real-world applications such as the LoD3 semantic reconstruction, we utilized the personalized SD-inpainting model to refine the incomplete real-world conflict maps.
%
% \subsubsection{Implementation of Other Baselines}

\section{Discussion}
% \subsection{Insights in General Performance}
% \thomasworries{Concerning randomly generated masks, FacaDiffy achieves an improvement of $+0.19$ in the SSIM, and of $-0.16$ in the LPIPS compared to the masked conflict map in Table \ref{tab:Results1}.}
%Concerning randomly generated masks, our experiments on the CMP-extended database demonstrate an improvement of $+0.19$ in the SSIM, and of $-0.16$ in the LPIPS compared to the masked conflict map in Table \ref{tab:Results1}.
FacaDiffy demonstrates enhancements of $+0.19$ in the SSIM and of $-0.16$ in the LPIPS compared to the masked conflict map in \autoref{tab:Results1}.
As illustrated in \autoref{fig:Exemplary tree results}, the pre-trained SD-inpainting model struggles to accurately reconstruct the facade structures from the tree-shaped inpainting mask.
This poses a challenge for deploying it in realistic scenarios, as trees represent a common type of occlusion. 
Conversely, our personalized model, FacaDiffy, effectively reduces the impact of tree-shaped masks and successfully completes the masked regions with semantically meaningful content.
This improvement is further substantiated in \autoref{tab:Results1} and \autoref{tab:ResultsAblation}, where FacaDiffy contributes to up to $10 \%$ boosts in the LPIPS scores compared to the pre-trained model.

%\subsection{Evaluation of the Results}
\subsection{Impact on the LoD3 Semantic Reconstruction}
 \begin{figure*}[tp]
     \centering
     \includegraphics[width=0.9\textwidth]{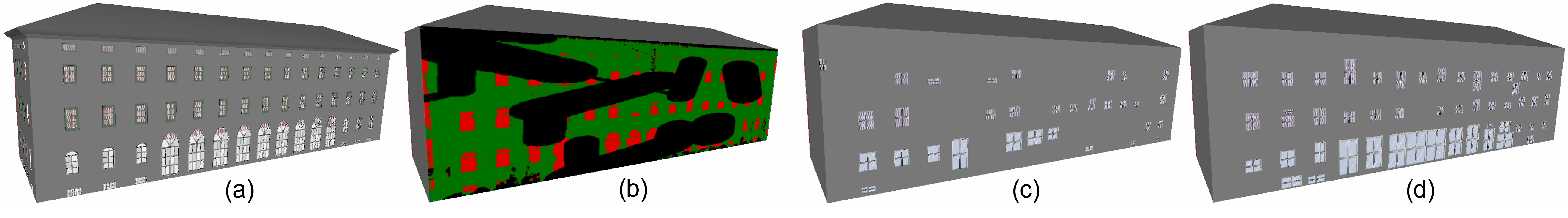}
     \caption{Impact of our method on the 3D reconstruction: (a) ground-truth LoD3 model; (b) incomplete conflict map; (c) 3D reconstruction based on the incomplete map; (d) 3D reconstruction based on the inpainted conflict map.}
     \label{fig:lod3reconstructionComparison}
     \vspace{0.1cm}
 \end{figure*}

%\olafworries{Olaf's bit:}
To evaluate FacaDiffy's impact on the potential downstream tasks, we selected one of the most prominent: \gls{LOD}3 reconstruction.
We selected building facades belonging to the so-called Building 23 of the TUM City Campus, characterized by three inpainting scenarios: A, very good facade visibility (visibility \textgreater90\%) with our randomly generated mask, B partial visibility due to scaffolding obstructing the field-of-view ($\sim$ 50\%), C low visibility due to trees and city furniture blocking the field-of-view (70\%).
We used the provided implementation for the part of the 3D reconstruction of \cite{wysocki2023scan2lod3}.
Yet, unlike the Scan2LoD3 probabilistic maps \cite{wysocki2023scan2lod3}, in the presented method we design our maps as hard evidence of conflict, confirmation, and unknown; additionally, we assumed that each conflict represented a window opening for homogeneous comparison.
We evaluated our method's impact for the typical three pillars of 3D semantic reconstruction in the following sub-sections.

\paragraph{Detection rate.}
Our experiments corroborate that FacaDiffy can achieve a higher detection rate by 22\% compared to the plain, deterministic approach (\autoref{tab:mofaDetection}).
The prominent example of the method's impact is illustrated in \autoref{fig:lod3reconstructionComparison}, where the 3D reconstruction of the randomly-masked facade A with (\autoref{fig:lod3reconstructionComparison}c) and without inpainting (\autoref{fig:lod3reconstructionComparison}d) is shown, indicating the 22\% higher detection rate.
Yet, the facade completeness remains challenging for small-size windows, as indicated by missing top-row windows in \autoref{fig:lod3reconstructionComparison}.
It is worth noting that FacaDiffy does not violate the main trait of the conflict maps: low false alarm rate. It remained negligible and similar throughout the experiments ($\sim$ 1\%).
\begin{table}[h]
\centering
%\small
\setlength\tabcolsep{1.8pt}
%\begin{tabular}{llllllllllllllllll}
\begin{tabular}{lcccccccccc}
%\begin{tabular}{lcccccccccccccccc}
\toprule
%\multirow{}{}{} & \multicolumn{4}{c}{H\&G \cite{tuttas_reconstruction_2013}} & \multicolumn{4}{c}{CC \cite{wysockiVisibility}} & \multicolumn{4}{c}{our (TUM)} & \multicolumn{4}{c}{our (MF)}  \\
& \multicolumn{4}{c}{Before} & \multicolumn{4}{c}{After} & \multicolumn{2}{c}{} \\
& \multicolumn{4}{c}{inpainting} & \multicolumn{4}{c}{inpainting} & \multicolumn{2}{c}{Gain $\uparrow$}  \\
\cmidrule(lr){2-5} \cmidrule(lr){6-9} \cmidrule(lr){10-11} % \cmidrule(lr){14-17}
 & A & B & C & Tot & A & B & C & Tot &  &  \\
 AO & 66 & 17 & 20 & \textbf{103} & 66 & 17 & 20 & \textbf{103} &  &  \\
\midrule
			 %MO & x & 17 & x & \textbf{x} & x & 17 & x & \textbf{x}  \\
			 D & 33 & 6 & 10 & \textbf{49} & 52 & 13 & 12 & \textbf{77} & $\uparrow$ & 28 \\
			 TP & 32 & 6 & 10 & \textbf{48} & 46 & 13 & 12 & \textbf{71} & $\uparrow$ & 23 \\
			 FP & 1 & 0 & 0 & \textbf{1} & 1 & 0 & 0 & \textbf{1} & $\sim$ & 0 \\
			 FN & 35 & 11 & 10 & \textbf{56} & 20 & 4 & 8 & \textbf{32} & $\uparrow$ & 24 \\%\hline
\midrule
%DA & 48.5\% & 35.3\% & 50\% & \textbf{46.6\%} & 69.7\% & 76.5\% & 60\% & \textbf{68.9\%} & $\uparrow$ & 22\% \\
DA & 48\% & 35\% & 50\% & \textbf{47\%} & 70\% & 76\% & 60\% & \textbf{69\%} & $\uparrow$ & 22\% \\
FA & 3\% & 0\% & 0\% & \textbf{2\%} & 2\% & 0\% & 0\% & \textbf{1\%} & $\uparrow$ & 1\% \\
%DM & x & 71 & x & \textbf{x} & x & 88 & x & \textbf{x} \\
%FR-MO & 0 & 0 & 50 & \textbf{4} & 0 & 0 & 16.7 & \textbf{1.2} & 0 & 0 & 0 & \textbf{0} & 0 & 12.5 & 6.2 & \textbf{3.1}\\
\bottomrule
\end{tabular}
    \caption{Detection rate for all openings (DA) and the respective false alarm rate (FA) for façades A, B, and C (AO = all openings, D = detections, TP = true positives, FP = false positives, FN = false negatives, $\uparrow$ indicates positive change).}
    \label{tab:mofaDetection}
\end{table} 
\paragraph{Segmentation.}
As we show in \autoref{tab:segmentationIoU}, the impact on the segmentation per-instance area was even across the testing sample.
We observe a decreasing performance rate correlated with the number of the initially detected windows (\autoref{tab:mofaDetection}). 
For example, the frontal inpainted facade A scored an improved \gls{IoU} over the non-inpainted one. 
Whereas a decrease was noticed for facade B owing to the small initially detected evidence sample (heavily obstructed by scaffolding); and a smaller decrease for facade C with a smaller obstruction rate.
Note, that here we did not include the samples with misses (FN).
\begin{table}[h]
\centering
%\small
\setlength\tabcolsep{4pt}
\begin{tabular}{lcccc}
\toprule 
%\multirow{}{}{} & \multicolumn{4}{c}{median IoU $\uparrow$}\\
& \multicolumn{4}{c}{mean IoU $\uparrow$}\\
\cmidrule(l){2-5}
Façade & A & B & C & Total \\
Openings & 66 & 17 & 20 & 103 \\
\midrule
% PT+Ft. \cite{zhao2021point}  & 7.3 & 4.6 & 3.7 & 7.3 \\
% M-RCNN \cite{he2017mask}  & 63.7 & 47.4 & 38.6 & 58.4 \\
% CC \cite{wysockiVisibility}  & 66.5 & 56.4 & \textbf{53.2} & 60.6 \\
% Scan2LoD3 (TUM)  & 63.9 & 52.9 & 38 & 62.1 \\
% Scan2LoD3 (MF)  & \textbf{78.4} & \textbf{62.3} & 40.6 & \textbf{76.2} \\
Before inpainting & 0.46 & 0.41 & 0.25 & 0.41 \\
After inpainting & 0.48 & 0.28 & 0.22 & 0.41 \\
\bottomrule
\end{tabular}
	\caption{Comparison of the median intersection over union (IoU) scores for the opening segmentation (openings where $IoU > 0\%$)}
\label{tab:segmentationIoU}
\end{table}
\vspace{-0.4cm}
\paragraph{3D reconstruction.}
Subsequently, we tested the impact on the final \gls{LOD}3 reconstruction for the complete building 23.
The Hausdorff distance was employed to calculate the difference between the ground truth \gls{LOD}3 model before and after the inpainting, which essentially evaluated the geometrical gain of the proposed method \cite{cignoni1998metro}. 
%For a more elaborate discussion on the conflict-based approaches advantages over other state-of-the-art methods, see \cite{wysocki2023scan2lod3}.
Here, the improvement oscillated around 0.03\si{m} in terms of the RMSE score and 0.01\si{m} for the  mean ($\mu$), as shown in \autoref{tab:Hausdorff}. 
As expected, the geometrical gain was relatively small, as the tested opening elements were not facade-protruded. 
\begin{table}[htb]
    %\small
    \centering
    \begin{tabular}{l ccc} 
	\toprule
	Method  & \multicolumn{2}{c}{vs. GT LoD3 $\downarrow$} \\
	\cmidrule(lr){2-3}
    & $\mu$ & RMS \\
    \midrule
    %Poisson~\cite{kazhdan2006poisson} & 0.35 & 0.54 & \xmark \\
    % CC~\cite{wysockiVisibility} & 0.31 & 0.34 & \cmark \\
    % Scan2LoD3 (TUM) & 0.23 & 0.26 & \cmark \\
    % Scan2LoD3 (MF)  & \textbf{0.13} & \textbf{0.25} & \cmark \\
    Before inpainting  & 0.27 & 0.30 \\
    After inpainting  & 0.26 & 0.27 \\
    \bottomrule
    \end{tabular}
    \caption{The impact of our method on the final LoD3 reconstruction using the ground-truth LoD3 model and the Hausdorff distance.}
    \label{tab:Hausdorff}
\end{table}

\subsection{Limitations}
Succeeding in effectively obtaining conflict maps, our deterministic ray analysis approach exhibits limitations in handling scenarios involving intricate transparent facade structures and buildings with extruded facade elements that encase corners. 
The lack of probabilistic information potentially limits its effectiveness in scenarios encompassing additional confidence information. 
The heuristically designed text prompt proves effective for the highly-challenging scenarios of random and tree occlusions. 
Yet, our method can be sensitive to the chosen text prompt. 
The degree to which the synthetically generated conflict maps resemble real conflict maps requires further investigation and is planned as future research.

\section{Conclusion}
We propose FacaDiffy, a method for completing conflict maps derived from semantic \gls{LOD}2 building models and corresponding laser scanning point clouds. 
% By personalizing with Dreambooth, we facilitate the deployment of a pre-trained \gls{SD} model to obtain completed conflict maps, outperforming traditional inpainting methods by up to $10\%$ in the \gls{LPIPS} score.
% Leveraging a stochastic city model generator to generate synthetic conflict maps makes it possible to complement insufficient real-world training data.
% 
FacaDiffy incorporates DreamBooth-based personalization to address the limitations of the pre-trained Stable Diffusion model in conflict map inpainting. 
This results in an approximate $10\%$ improvement in the \gls{LPIPS} score when dealing with tree-shaped inpainting masks.
Interestingly, as illustrated in Figure \ref{fig:counterintuitive_Evaluation}, the \gls{IoU} exhibits counterintuitive evaluation results.
Extensive experiments further demonstrate the superior performance of our method over other image inpainting baselines. 
% FacaDiffy 
% % personalizing with Dreambooth 
% mitigates the weakness of the pre-trained Stable Diffusion model concerning tree-shaped masks by up to $10\%$ in the \gls{LPIPS} score compared to the pre-trained Stable Diffusion model.
During the \gls{LOD}3 semantic reconstruction, the introduction of FacaDiffy contributes to a $22\%$ increase in the detection rate, proving its efficacy in enhancing the 3D reconstruction accuracy of existing pipelines. 
% In addition, FacaDiffy achieves an increase in the detection rate of $22\%$, proving effective in enhancing the 3D reconstruction accuracy of existing pipelines. 
Future work will explore the completed conflict maps in the various application pipelines, such as solar potential estimation of facades and wind flow simulations.

\section*{Acknowledgements}

This work was conducted within the framework of the Leonhard Obermeyer Center at the Technical University of Munich (TUM). 
We thank the City of Munich for the cooperation in the Connected Urban Twins (CUT) project funded by the Federal Ministry for Housing, Urban Development and Building of Germany (BMWSB).
We gratefully acknowledge the staff members of the TUM Professorship of Photogrammetry and Remote Sensing for their valuable insights and support.

{
	\begin{spacing}{1.03}
		\normalsize
		\bibliography{bibliography} % Include your own bibliography (*.bib), style is given in isprs.cls
	\end{spacing}
}
\end{document}